\documentclass[10pt,twocolumn,letterpaper]{article}

\usepackage{iccv}
\usepackage{times}
\usepackage{epsfig}
\usepackage{graphicx}
\usepackage{amsmath}
\usepackage{amssymb}
\usepackage{verbatim}
\usepackage{tabularx,booktabs}

\usepackage[pagebackref=true,breaklinks=true,letterpaper=true,colorlinks,bookmarks=false]{hyperref}

\iccvfinalcopy 

\def\iccvPaperID{3385} 

\ificcvfinal\fi
\begin{document}

\title{FaceShapeGene: A Disentangled Shape Representation \\ for Flexible Face Image Editing}

\author{Sen-Zhe Xu\textsuperscript{\dag\ddag}\thanks{Work done while Sen-Zhe Xu was a Research Intern with Tencent AI Lab.}, Hao-Zhi Huang\textsuperscript{\ddag}, Shi-Min Hu\textsuperscript{\dag}, Wei Liu\textsuperscript{\ddag}\\
\textsuperscript{\dag}Tsinghua University \textsuperscript{\ddag}Tencent AI Lab\\
 \tt\small{xsz15@mails.tsinghua.edu.cn, huanghz08@gmail.com, shimin@tsinghua.edu.cn, wl2223@columbia.edu}
}

\maketitle

\begin{abstract}
Existing methods for face image manipulation generally focus on editing the expression, changing some predefined attributes, or applying different filters. However, users lack the flexibility of controlling the shapes of different semantic facial parts in the generated face.
In this paper, we propose an approach to compute a disentangled shape representation for a face image, namely the \emph{FaceShapeGene}. The proposed \emph{FaceShapeGene} encodes the shape information of each semantic facial part separately into a 1D latent vector. 
On the basis of the \emph{FaceShapeGene}, a novel part-wise face image editing system is developed, which contains a shape-remix network and a conditional label-to-face transformer.
The shape-remix network can freely recombine the part-wise latent vectors from different individuals, producing a remixed face shape in the form of a label map, which contains the facial characteristics of multiple subjects. The conditional label-to-face transformer, which is trained in an unsupervised cyclic manner, performs part-wise face editing while preserving the original identity of the subject.
Experimental results on several tasks demonstrate that the proposed \emph{FaceShapeGene} representation correctly disentangles the shape features of different semantic parts.
Comparisons to existing methods demonstrate the superiority of the proposed method on accomplishing novel face editing tasks. 

\end{abstract}

\begin{figure}[t]
    \centering
    \includegraphics[width=1.0\linewidth]{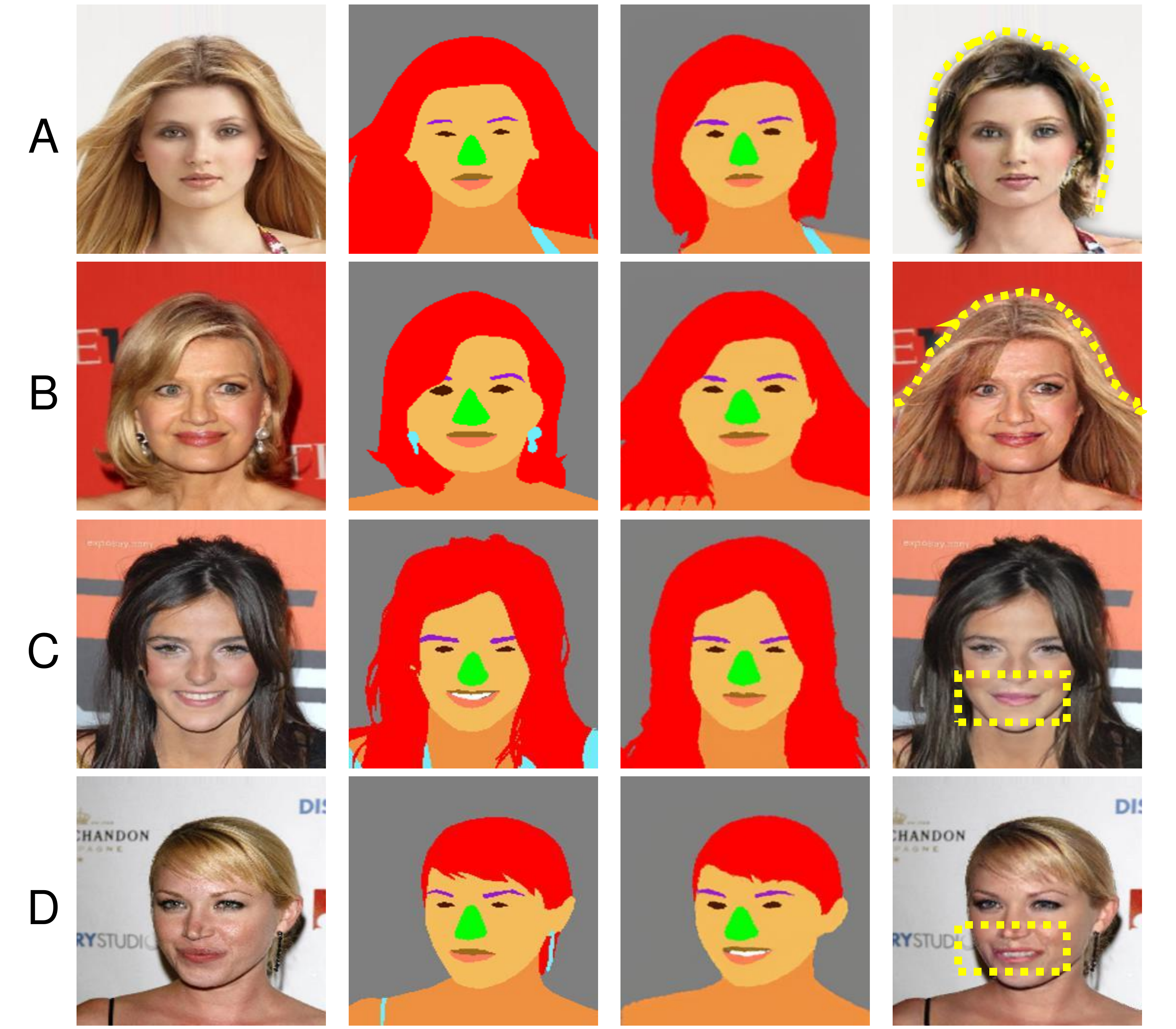}
    \caption{Partial face editing with \emph{FaceShapeGene}. The first column shows the input face images. The second column shows the corresponding label maps. The third column are the remixed label maps generated by our shape-remix network. The fourth column are the generated face images by our conditional label-to-face transformer. Note that we exchange the hair styles between A and B, and exchange the mouths between C and D.}
    \label{fig:intro}
    \vspace{-5pt}
\end{figure}

\vspace{-5pt}
\section{Introduction}
\vspace{-2pt}
With the rapid development of deep generative models, image generation results evolve to be more and more realistic. Face is one of the most significant image categories that people care about, due to its wide applications in various aspects. The leading technologies for image generation include Generative Adversarial Networks (GANs)~\cite{goodfellow2014generative, berthelot2017began, arjovsky2017wasserstein, gulrajani2017improved, karras2017progressive, li2018unsupervised}, Variational Auto-Encoders (VAEs)~\cite{kingma2013auto,rezende2014stochastic,larsen2015autoencoding, higgins2016beta,mescheder2017adversarial, bao2017cvae}, and Flow-based Generative Models (FGMs)~\cite{dinh2014nice, dinh2016density, kingma2018glow}. Among them, GANs receive most of the attentions and implicate a lot of variants that enable a variety of face generation applications.


Besides randomly synthesizing a realistic face image based on a latent vector~\cite{goodfellow2014generative, berthelot2017began, arjovsky2017wasserstein, gulrajani2017improved, karras2017progressive}, people are more interested in developing approaches to support more creative and flexible face editing tools. Transforming a label map or a sketch into a corresponding face image can be formulated as an image-to-image translation problem~\cite{isola2017image, zhu2017unpaired, wang2018pix2pixHD}. By editing the label map or the sketch manually, we can generate face images with different overall layouts. However, the identity of the generated person is out of control in this scenario. 
By training with an auxiliary classification loss, the image-to-image translation technique can also be applied to manipulating one or multiple predefined attributes of a face image~\cite{choi2018stargan, zhao2018modular}, such as the gender, age, expression and hair color. GANimation~\cite{pumarola2018ganimation} can generate arbitrary facial expressions by training a GAN model conditioned on a continuous embedding of muscle movements. When applying the above mentioned methods, the face identity is preserved through a reconstruction loss. Another kind of identity-preserved face editing methods focus on view synthesis~\cite{huang2017beyond, shen2018faceid}. They rely on a pretrained face recognition model to compute an identity loss to constrain the model training.

However, the above methods have several drawbacks. Firstly, we cannot edit a certain facial part without touching the others. Secondly,  the number of available editing operations are constrained by the dataset. For example, CelebA~\cite{liu2015faceattributes} only contains 40 types of binary attributes. Thirdly, when given a reference image, we cannot conveniently transfer the attribute of a reference image to our target image.

In this paper, we propose a novel disentangled shape representation, namely \emph{FaceShapeGene}, which supports flexible face editing manipulations. Specifically, based on a large-scale face parsing dataset, we train multiple local face parsers to extract the shape features of different semantic parts. 
The semantic parts labeled in the dataset include hair, eyebrows, eyes, nose, mouth, face shape and the upper body. 
This face parsing dataset will be made publicly available in the future.
On the basis of the disentangled shape representation, we propose a shape-remix network to freely recombine the latent vectors of multiple facial parts belonging to different individuals to produce a remixed \emph{FaceShapeGene}. 
We can decode the remixed \emph{FaceShapeGene} into a remixed label map, which contains the local shape features from different faces. Meanwhile, we propose a conditional label-to-face transformer which takes the remixed label map and a conditional face as input, generating a remixed face. The generated remixed face preserves the identity of the conditional input. The proposed conditional label-to-face transformer is trained in an unsupervised manner with adversarial losses and cycle-consistent losses in both the label and image domains. By coupling the shape-remix network and the conditional label-to-face transformer together, we can modify a desired part of a given face to make it resemble the part of another face, while keeping the original identity. Thus, the available facial attributes are no longer restricted to the ones provided by CelebA or other dataset. We test our system on several novel face editing applications, including exchanging the hair style between two individuals,  manipulating the smile, redrawing the eyebrows, resizing the nose and so on. An example is shown in Fig.~\ref{fig:intro}. Extensive experiments have been conducted to demonstrate the necessity of each component of our method.



In summary, the main contributions of this paper are four-fold. Firstly, we present a method for extracting a disentangled shape representation of a face image. Secondly, we propose a shape-remix network to recombine the shape features of different individuals to easily generate a desired label map. Thirdly, we propose a conditional label-to-face transformer to generate a face image according to the given label map while preserving the original identity. Fourthly, some novel face editing manipulations are developed by coupling the shape-remix network and the conditional label-to-face transformer together.

\vspace{-5pt}
\section{Related Work} \label{sec:related}
\vspace{-2pt}

\paragraph{General Image-to-Image Translation.}
Image-to-image translation is a problem of translating one possible representation of an image into another representation.
Isola \etal~\cite{isola2017image} proposed Pix2Pix to give a supervised solution to general image-to-image translation based on conditional adversarial networks.
Afterwards, some unsupervised methods~\cite{zhu2017unpaired, liu2017unsupervised} were proposed by introducing the cycle-consistent constraints. The above methods make a simplifying assumption that image-to-image translation is a problem of learning a deterministic one-to-one mapping. However, one-to-many or many-to-many mappings exist in most of the image-to-image translation tasks. Recently, methods like MUNIT~\cite{huang2018munit} and DRIT~\cite{DRIT2018} tried to tackle the multimodal image-to-image translation problem by decomposing the latent representation of an image into a domain-invariant content code and a domain-specific style code, greatly reducing mode collapse and producing diverse multimodal translation results. 
There are also some approaches that intend to perform image-to-image translation at higher resolution, under either a supervised setting~\cite{wang2018pix2pixHD} or an unsupervised setting~\cite{li2018unsupervised}.

\begin{figure*}[t]
    \centering
    \includegraphics[width=0.9\linewidth]{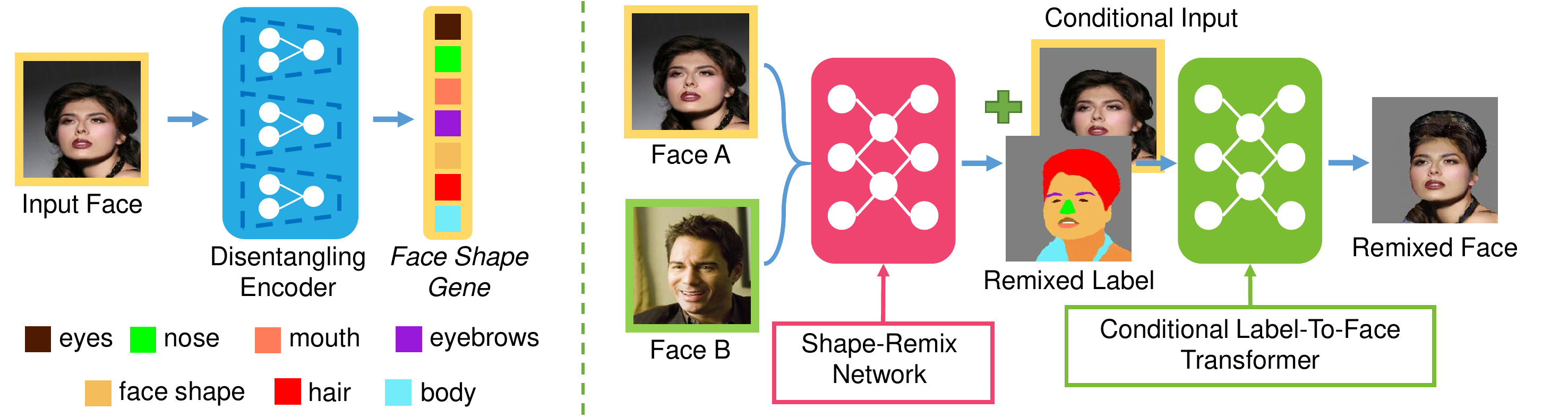}
    \caption{The overall pipeline of our system.
    A disentangling encoder is trained to extract the proposed \emph{FaceShapeGene} representation, which divides the shape information of the whole face into seven semantic parts. Based on this representation, a shape-remix network is proposed to produce a remixed face label map by recombining the \emph{FaceShapeGene}s of multiple faces. In addition, a conditional label-to-face transformer is proposed to transform the remixed label map into a corresponding photo-realistic face image while preserving the identity of the conditional input. }
    \label{fig:pipeline}
    \vspace{-5pt}
\end{figure*}

\vspace{-10pt}
\paragraph{Face Editing with GANs.}
Besides studying the general image-to-image translation problem, a large number of GANs focus on face editing. GANnimation, proposed by Pumarola \etal~\cite{pumarola2018ganimation}, delves deeper into the expression editing task and utilizes the Facial Action Coding System (FACS) for describing a continuous embedding of muscle movements. By training a GAN model conditioned on the FACS codes, GANnimation can manipulate the expression of the face images. Choi \etal proposed StarGAN~\cite{choi2018stargan} to integrate multiple two-domain facial attribute transformations in a single model. Zhao \etal proposed ModularGAN~\cite{zhao2018modular} to accomplish editing multiple facial attributes at the same time by stacking multiple two-domain models in series.
Besides transforming some predefined facial attributes, there are also some methods working on generating faces from arbitrary viewpoints, while preserving the identity. Shen \etal~\cite{shen2018faceid} proposed a three-player GAN, which considers pose, identity, and realism simultaneously. In this paper, we propose a novel disentangled shape representation for face images, which is the basis of a novel part-aware face editing system. Our approach can also preserve the identity of the input face by utilizing cycle-consistent losses. While previous approaches~\cite{pumarola2018ganimation, choi2018stargan, zhao2018modular} merely employ a cycle-consistent loss in the image domain, we also introduce a cycle-consistent loss in the label domain to ensure that the shape-remix network works as expected.


\vspace{-5pt}
\section{Approach} \label{sec:method}
\vspace{-2pt}

We propose \emph{FaceShapeGene}, a disentangled shape representation which benefits the development of flexible face editing tools. The overall pipeline is shown in Fig.~\ref{fig:pipeline}. The proposed \emph{FaceShapeGene} encodes the shape information of different facial parts into 1D latent vectors, respectively. 
On the basis of \emph{FaceShapeGene}, a part-wise face editing system is developed. Our proposed system consists of two components. The first component is a shape-remix network, which recombinies the \emph{FaceShapeGene}s from multiple faces and produces a remixed face shape in the form of a label map. Exploiting the proposed shape-remix network, we can conveniently transfer the shape of a facial part from a reference image to a target image. The second component is a conditional label-to-face transformer, which takes the remixed label map and a conditional face image as input to generate a remixed face image. While the remixed label map provides the desired shape information, the conditional face image offers the identity information we want to keep. The shape-remix network and the conditional label-to-face transformer are coupled together and trained in an unsupervised cyclic manner to learn to generate a realistic face image in the desired shape with the target identity. 

\begin{figure}[t]
    \centering
    \includegraphics[width=0.8\linewidth]{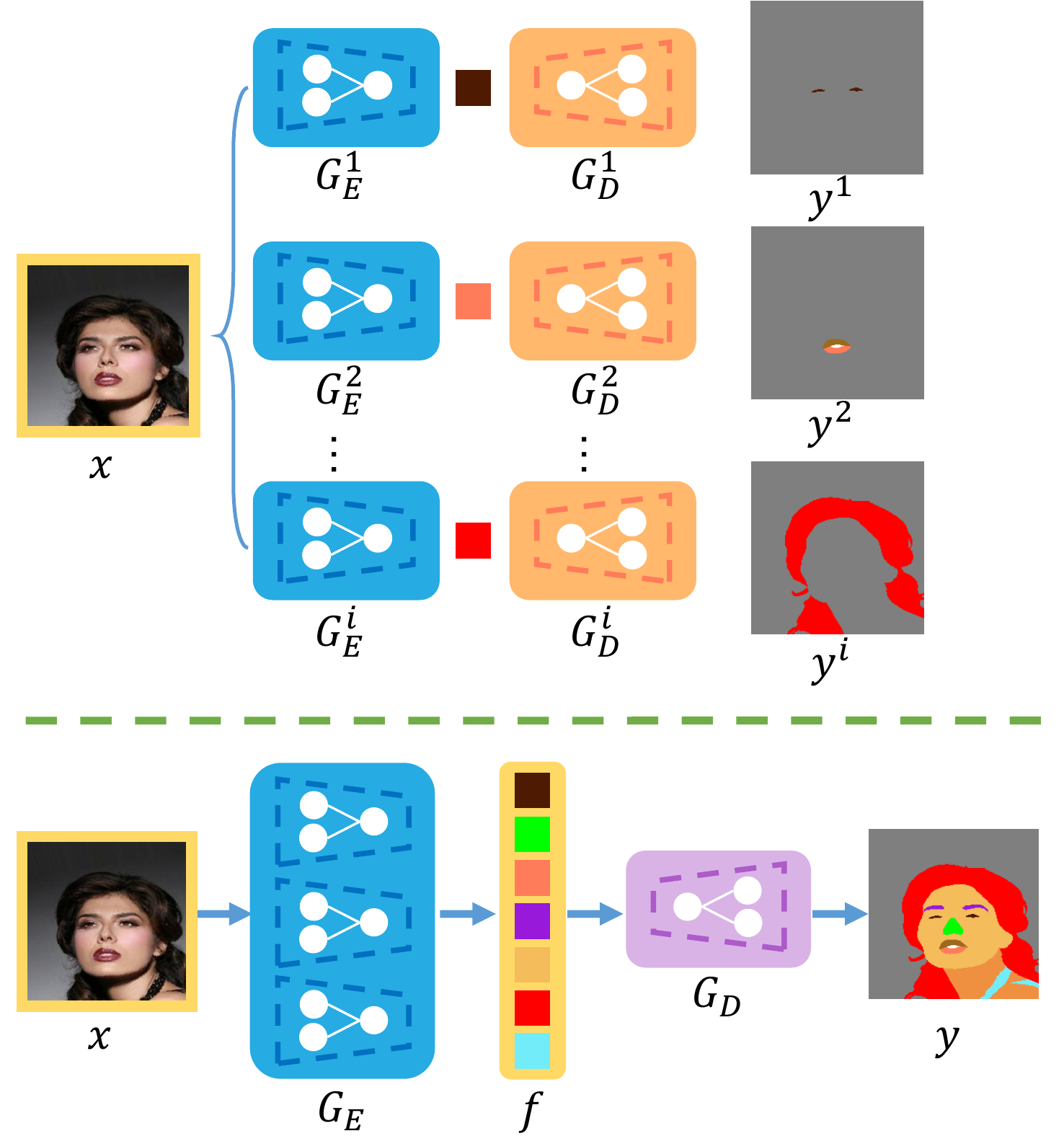}
    \caption{Learning disentangled shape representations. 
    An individual local face parser $(G^{i}_{E}, G^{i}_{D})$ is trained for the $i$-th facial part to extract part-wise shape information in the form of 1D latent vector. All the part-wise shape features are concatenated together to form the \emph{FaceShapeGene} $f$. Finally, an overall decoder $G_{D}$ is trained to decode $f$ into a whole-face label map $y$.}
    \label{fig:learning_faceshapegene}
    \vspace{-15pt}
\end{figure}

\vspace{-2pt}
\subsection{Learning The FaceShapeGene}
We propose a disentangled shape representation \emph{FaceShapeGene} for face images, with which we can modify the shape of each facial part individually by editing the corresponding part-wise feature. 
In this paper, we focus on seven facial parts, including hair, eyebrows, eyes, nose, mouth, face shape, and body.
As shown in Fig.~\ref{fig:learning_faceshapegene}, the shape information of different parts are disentangled by training an individual local face parser $(G^{i}_{E}, G^{i}_{D})$ for each facial part. For the $i$-th facial part, an encoder $G^{i}_{E}$ encodes $x$ into a 1D latent vector, while a decoder $G^{i}_{D}$ decodes the 1D latent vector into a partial label map $y^i$. 
Our label map is a three-channel image in the RGB color space, in which each facial part is represented by a unique color, except for the mouth and body. For the mouth, in order to increase the representative capacity, we use three different colors to denote the upper lip, lower lip, and teeth. For the body, we use two different colors for the body skin and clothes.
Since the supervision signal is a label map, which contains no texture and color of the original face, the local face parser will automatically discard the detailed appearance information during training.


We formulate the local face parsing task as a regression problem rather than a dense classification problem. This is because when formulating the label map prediction task as a dense classification problem using the cross-entropy loss as in the scene parsing task~\cite{chen2018deeplab, chen2017rethinking, chen2018encoder}, the minority category with fewer pixels, like the eyes, tends to be neglected. On the contrary, formulating the task as a regression problem can alleviate the above class imbalance issue. To train the local face parsers, a combination of an $L_1$ loss, a VGG loss and a GAN loss is adopted. The  $L_1$ loss is a pixel-wise reconstruction loss defined as:
\begin{small}
\begin{equation}
\begin{aligned}
\small{
\mathcal{L}_{L_1}(G^{i}_{E},G^{i}_{D})= \mathbb{E}_{x,y}\left[\left\|G^{i}_{D}(G^{i}_{E}(x))-y^{i} \right\|_1 \right],
}
\end{aligned}
\end{equation}
\end{small}
where $x$ denotes the input image,  $y^{i}$ denotes the ground-truth partial label map corresponding to the $i$-th part. 
However, in our experiments, using only an $L_1$ loss leads to blurry generated label maps. To alleviate this issue, a VGG loss and a GAN loss are introduced. The VGG loss is a perceptual reconstruction loss~\cite{Johnson2016Perceptual, ledig2017photo} defined as:
\begin{small}
\begin{equation}
\begin{aligned}
\small
\mathcal{L}_\text{VGG}(G^{i}_{E},G^{i}_{D})= \mathbb{E}_{x,y}\left[\left\|\phi(G^{i}_{D}(G^{i}_{E}(x)))-\phi(y^{i}) \right\|_1 \right],
\end{aligned}
\end{equation}
\end{small}
where $\phi$ denotes the features extracted by a pretained VGG19 network~\cite{simonyan2014very}. In our experiments, we use the \emph{conv}1-1, \emph{conv}2-2, \emph{conv}3-2, \emph{conv}4-4, and \emph{conv}5-4 layers to extract features.
The VGG loss encourages the generated result to be similar to the ground-truth in the semantic feature domain.
We also adopt a discriminator $D^{i}$ for the $i$-th facial part to calculate the GAN loss:
\begin{small}
\begin{equation}
\begin{aligned}
\small
\mathcal{L}_\text{GAN}(G^{i}_{E},G^{i}_{D},D^{i}) &= \mathbb{E}_{x,y}\left[ \| D^{i}(x, y^i) \|_2^2  \right] \\
& + \mathbb{E}_{x}\left[ \| 1- D^{i}(x, G^{i}_{D}(G^{i}_{E}(x))) \|_2^2  \right].
\end{aligned}
\end{equation}
\end{small}
Here, we use LSGAN~\cite{mao2017least} and PatchGAN~\cite{isola2017image} for stable training. $\mathcal{L}_\text{GAN}$ ensures that the generated partial label map $\hat{y}^i$ stays in the label domain, which brings more perceptual details.
The full objective function for training the local face parser {\small$(G^{i}_{E},G^{i}_{D})$} is: {\small$\mathcal{L}_\text{total}(G^{i}_{E},G^{i}_{D}, D^i) = \mathcal{L}_\text{$L_1$} + \lambda_\text{VGG}\mathcal{L}_\text{VGG} + \lambda_\text{GAN}\mathcal{L}_\text{GAN}$}.
We optimize the objective function by alternately updating the local face parser {\small$(G^{i}_{E},G^{i}_{D})$} and the discriminator {\small$D^i$}:
{\small$\arg\min_{G^{i}_{E},G^{i}_{D}}\max_{D^{i}}\mathcal{L}_\text{total}(G^{i}_{E},G^{i}_{D}, D^i)$}.

Once the training of all the local face parsers is done, we train an overall decoder $G_{D}$ to gather the shape information of the whole face, as shown at the bottom of Fig.~\ref{fig:learning_faceshapegene}. We name the collection of all the part-wise encoders as a disentangling encoder $G_E$, which concatenates all the part-wise 1D latent vectors to formulate an overall 1D latent vector $f$. Then, $f$ is fed to the overall decoder $G_{D}$ to produce the final whole-face label map $y$. Similarly, we use a combination of $L_1$, VGG, and GAN losses to train $G_{D}$. $G_E$ is fixed during training $G_D$.

\begin{figure}[t]
    \centering
    \includegraphics[width = 0.9\linewidth]{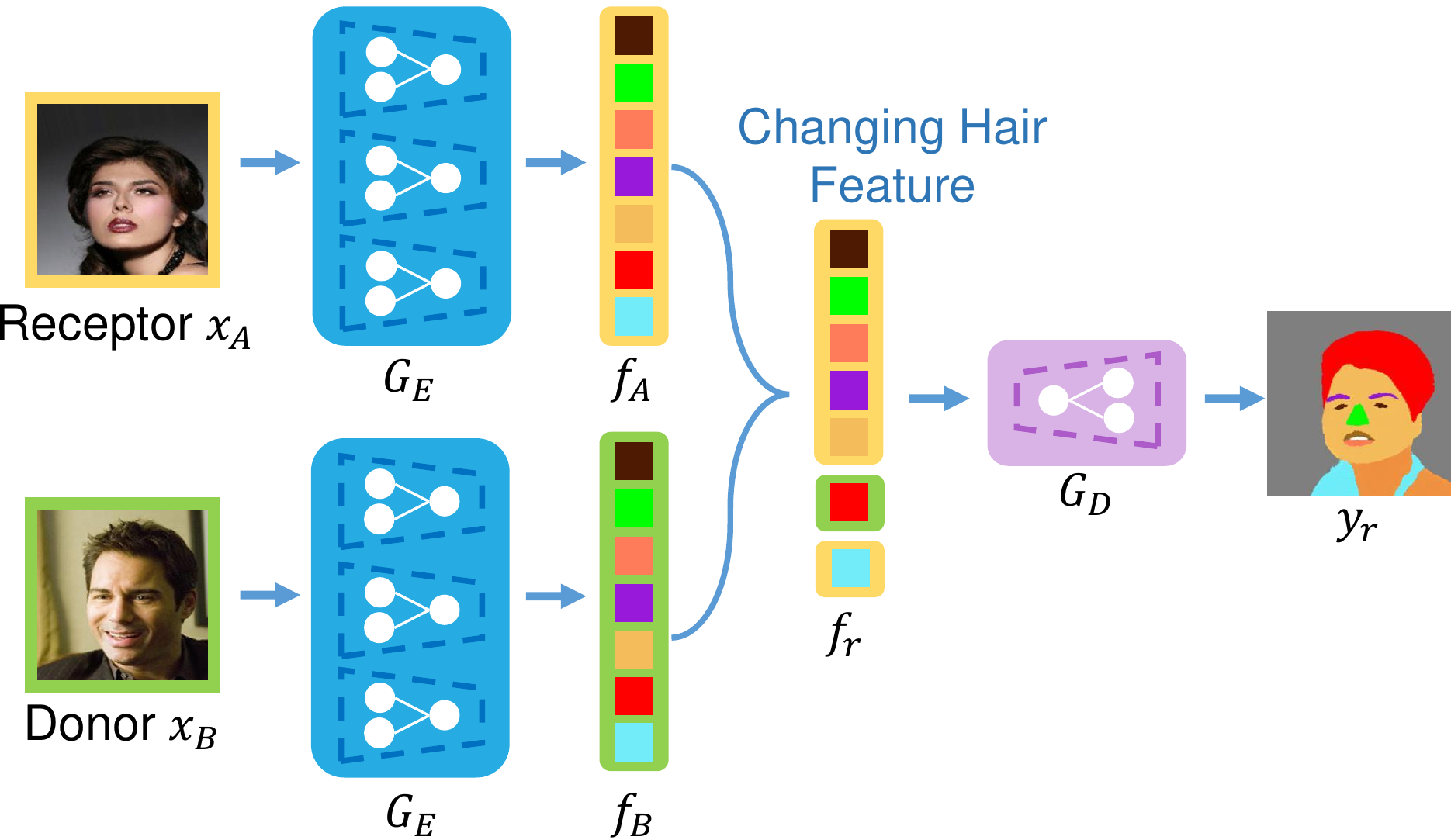}
    \caption{The shape-remix network. The shape-remix network employs a pretrained disentangling encoder $G_E$ to extract the \emph{FaceShapeGene}s   $f_A$ and $f_B$  for the receptor $x_A$ and the donor $x_B$, respectively. 
    By replacing the \emph{hair gene} of $f_A$ with that of $f_B$, a remixed \emph{FaceShapeGene} $f_r$ is obtained. Then, $f_r$ is decoded into a remixed label map $y_r$ by the overall decoder $G_D$. The same operation can be applied to other semantic parts.}
    \label{fig:shape_remix}
    \vspace{-5pt}
\end{figure}

\begin{figure*}[t]
    \centering
    \includegraphics[width=0.8\linewidth]{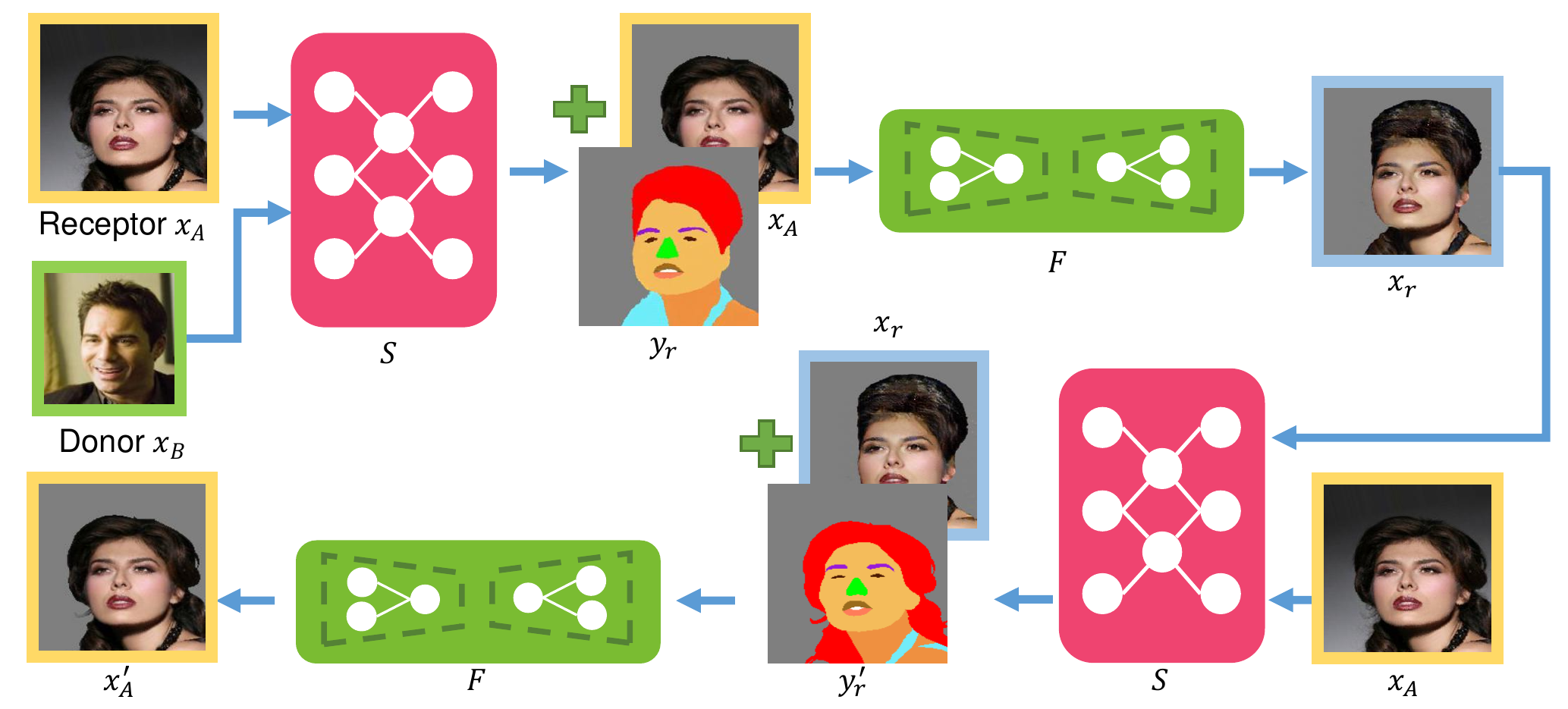}
    \caption{The cyclic training framework. The shape-remix network $S$ takes a pair of images $(x_A,x_B)$, producing a remixed whole-face label map $y_r$, by transferring the $i$-th part-wise shape feature of $x_B$ to $x_A$. The conditional label-to-face transformer $F$ takes $y_r$ and $x_A$ to generate a remixed face $x_r$. Symmetrically, the remixed face image $x_r$ and the receptor image $x_A$ are fed back to $S$ to reconstruct a remixed label map $y_r'$. $F$ takes $y_r'$ and the remixed face $x_r$ as input, generating a reconstructed image $x_A'$.
    }
    \label{fig:cyclic_training}
    \vspace{-5pt}
\end{figure*}

\vspace{-2pt}
\subsection{The Shape-Remix Network}
Based on the disentangled shape representation, we propose a shape-remix network $S$ to recombine the part-wise shape features from different people, generating a new whole-face label map containing the shape characteristics of multiple faces. This operation resembles the genetic editing process.
As shown in Fig.~\ref{fig:shape_remix}, an input image $x_A$ is treated as a receptor and a reference image $x_B$ acts as a donor. 
For both the receptor $x_A$ and the donor $x_B$, we employ the disentangling encoder $G_E$ to extract two \emph{FaceShapeGene}s $f_A$ and $f_B$. By replacing the \emph{hair gene} of $f_A$ with that of $f_B$, we obtain a remixed \emph{FaceShapeGene} $f_r$. Then, $f_r$ is decoded into a remixed whole-face label map $y_r$ by the overall decoder $G_D$. The same operation can be applied to other facial parts.



However, learning to generate the remixed whole-face label map $y_r$ corresponding to an arbitrary remixed \emph{FaceShapeGene} $f_r$ is still a challenge, because the overall decoder $G_{D}$ has not seen any remixed shape representation. Given the fact that there is no ground truth for $y_r$, supervised training for $G_{D}$ over the remixed shape representations is not possible. In the following section, we will introduce a cyclic training strategy coupled with a conditional label-to-face transformer to address the unsupervised fine-tuning of $G_{D}$.

\vspace{-2pt}
\subsection{The Conditional Label-to-Face Network}

On the basis of the shape-remix network $S$, we propose a conditional label-to-face transformer $F$ to complete the task of the part-wise shape editing of a face image. The transformer $F$ learns to transform a whole-face label map to a photo-realistic face image, while preserving the identity of a conditional input image. Since there is no ground-truth data for the supervised training of $F$, we adopt both adversarial losses and cycle-consistent losses to train $F$ in an unsupervised manner. The complete training process is illustrated in Fig.~\ref{fig:cyclic_training}. At the beginning, the shape-remix network $S$ takes a pair of images $(x_A,x_B)$ and produces a remixed whole-face label map $y_r=S(x_A,x_B,i)$. Here, $i$ indicates that the index of the targeted facial part. $i$ is randomly chosen for each iteration. Then, the transformer $F$ takes $y_r$ and $x_A$ as input, generating a remixed face $x_r=F(y_r,x_A)$.  Here, $x_A$ is a conditional input to provide the identity information.
In our experiments, we remove the background of the conditional input for more stable training. 
Note that from the inconsistency between $y_r$ and $x_A$, the transformer $F$ can implicitly identify the targeted part. The remixed face $x_r$ is supposed to be corresponding to the remixed label map $y_r$, while preserving the identity of the conditional input $x_A$.
Currently, the color or texture of the targeted facial part is generated randomly, because we only constrain the shape. The strategy for assigning the desired texture will be left for the future work.
Symmetrically, the remixed face image $x_r$ and the receptor image $x_A$ are fed back to the shape-remix network to reconstruct a remixed label map $y_r' =S(x_r,x_A,i)$. Note that $y_r'$ is supposed to be the same as $y_A$, which is the ground-truth label map for $x_A$. Next, $F$ takes the reconstructed label map $y_r'$ and the remixed face $x_r$ as input, generating a reconstructed image $x_A'$. Here, $x_A'$ is supposed to be the same as the input $x_A$.

\vspace{-10pt}
\paragraph{Cycle-Consistent Losses.} According to the above process, two cycle-consistent losses can be defined. The first one is in the label domain:
\begin{small}
\begin{equation}
\begin{aligned}
\mathcal{L}_\text{cyc,L} &=  \mathbb{E}_{y_r',y_A}\left[\left\| y_r' - y_A\right\|_1 + \lambda_\text{VGG} \left\|\phi(y_r')-\phi(y_A) \right\|_1 \right].
\label{eq:L_cyc_L}
\end{aligned}
\end{equation}
\end{small}
On one hand, $\mathcal{L}_\text{cyc,L}$ provides a supervision signal to finetune the overall decoder $G_D$ in the shape-remix network $S$, when regarding the remixed face $x_r$ as a receptor and the original receptor face $x_A$ as a donor, respectively. On the other hand, since the shape features except for the $i$-th part are provided by $x_r$, $\mathcal{L}_\text{cyc,L}$ encourages the remixed face $x_r$ to preserve the correct part-wise shapes to reconstruct a label map $y_r'$ which resembles $y_A$.
The second cycle-consistent loss is in the image domain:
\begin{small}
\begin{equation}
\begin{aligned}
\mathcal{L}_\text{cyc,I} &=  \mathbb{E}_{x_A',x_A}\left[\left\| x_A' - x_A \right\|_1 + \lambda_\text{VGG} \left\|\phi(x_A')-\phi(x_A) \right\|_1  \right].
\label{eq:L_cyc_I}
\end{aligned}
\end{equation}
\end{small}
Here, $x_A'=F(y_r',x_r)$ is a reconstructed face, which should be similar to the receptor face $x_A$. Note that the background of $x_A$ is also removed when computing $\mathcal{L}_\text{cyc,I}$.  It is reasonable to assume that $y_r'$ resembles $y_A$, given the cycle-consistent loss in the label domain. Thus, $\mathcal{L}_\text{cyc,I}$ provides a supervised ground-truth to train $F$ to generate a photo-realistic image $x_A'$ which is corresponding to the label map $y_A$. In addition, $\mathcal{L}_\text{cyc,I}$ also encourages the remixed face $x_r$ to provide correct identity information when acting as a conditional input. If $x_r$ does not preserve the identity of $x_A$, it cannot provide the correct identity information for reconstructing face $x_A'$.

\vspace{-10pt}
\paragraph{Adversarial Losses.} To ensure that $F$ is learning to generate a photo-realistic face image, an adversarial loss in the image domain is added:
\begin{small}
\begin{equation}
\begin{aligned}
\mathcal{L}_\text{GAN,I} &= \mathbb{E}_{x_A,x_B} \left[ \| D_I(x_A) \|_2^2 +  \| D_I(x_B) \|_2^2 \right] \\
& + \mathbb{E}_{x_r} \left[ \| 1 - D_I(x_r) \|_2^2 \right] \\
& + \mathbb{E}_{x_A'} \left[ \| 1 - D_I(x_A') \|_2^2 \right],
\end{aligned}
\end{equation}
\end{small}
where $x_r=F(y_r,x_A)$ and $x_A'=F(y_r',x_r)$ . 
Note that the overall decoder $G_{D}$ also undergoes finetuning during the training process. Thus, we need to make sure that $G_{D}$ still generates a meaningful label map by introducing a label-domain adversarial loss $\mathcal{L}_\text{GAN,L}$ similarly.

\vspace{-10pt}
\paragraph{Identity Constraint.} Since the traditional label-to-face transformation is a one-to-many mapping, we may have multiple face images corresponding to the same label map. The role of the conditional input image is to offer the identity information. The cycle-consistent losses mentioned above have already implicitly encouraged the transformer $F$ to learn the identity preservation. In order to explicitly constrain the identity, we propose a masked identity loss:
\begin{small}
\begin{equation}
\begin{aligned}
\mathcal{L}_\text{ID} &= \mathbb{E}_{x_r,x_A}\left[\left\| (x_r - x_A) \odot (1 - M_1) \right\|_1 \right] \\
&+ \mathbb{E}_{x_A',x_r'}\left[ \left\| (x_A' - x_r') \odot (1 - M_2) \right\|_1 \right].
\end{aligned}
\end{equation}
\end{small}
Here, $M_1$ and $M_2$ are editing masks computed by combining the $i$-th part-wise masks of both the receptor and donor, indicating the potential editing area. This loss term encourages that the content outside the editing area remains untouched.

\vspace{-10pt}
\paragraph{The Final Objective.} The final objective function is:
\begin{small}
\begin{equation}
\begin{aligned}
\mathcal{L}_\text{total} &= \mathcal{L}_\text{cyc,I} + \lambda_\text{CL} \mathcal{L}_\text{cyc,L}  + \lambda_\text{GI} \mathcal{L}_\text{GAN,I}  + \lambda_\text{GL} \mathcal{L}_\text{GAN,L} + \lambda_\text{ID} \mathcal{L}_\text{ID},
\end{aligned}
\end{equation}
\end{small}
which can be optimized in the form of a min-max game: $\arg \min_{F,G_D} \max_{D_I,D_L} \mathcal{L}_\text{total}(F,G_D,D_I,D_L)$. 



\vspace{-5pt}
\section{Experiments} \label{sec:experiment}
\vspace{-3pt}

\paragraph{Dataset.}
An internal face parsing dataset, containing 17,975 face images with ground-truth label maps, is utilized to train and test our network. This dataset will be made publicly available. The face images in this dataset are from CelebA~\cite{liu2015faceattributes}.  For each image, we hire people to manually draw a pixel-wise label map, while each part is labeled with a unique color. The specific parts include hair, eyebrows, eyes, nose, upper lip, lower lip, teeth, face skin, body skin, clothes, and background.  We split the dataset into three parts: 14,403 pairs for training, 1,781 pairs for validation, and 1,791 pairs for testing.

\vspace{-10pt}
\paragraph{Implementation Details}
During training, all the images and label maps are resized to $256\times 256$. A random affine transformation is adopted for data augmentation.
We adapt the architecture proposed by Johnson \etal~\cite{Johnson2016Perceptual} to build our system. Our part-wise encoder consists of 3 convolutional layers, 4 residual blocks~\cite{he2016deep}, and 2 additional convolutional layers to generate a feature map of size $128 \times 1 \times 1$.
Our part-wise decoder or the overall decoder consists of 2 transposed convolutional layers, 5 residual blocks, 2 more transposed convolutional layers, and one additional convolutional layer in the end to transform the feature map to a three-channel RGB label map. 
The architecture of $F$ consists of 5 convolutional layers, 9 residual blocks, and 4 transposed convolutional layers.
Please refer to the appendix for architecture details.
The Adam solver~\cite{kingma2014adam} with lr = $0.0002$ and $(\beta_1,\beta_2) = (0.5,0.999)$ is adopted. To learn the disentangled shape representation, we train the part-wise encoder-decoders for 100 epochs, and the overall decoder for 50 epochs. During the cyclic training of the whole system, we fix the disentangling encoder and train all the other networks together for 15 epochs. The best snapshot for each component is recorded according to the validation performance. When conducting the adversarial training, we adopt the gradient penalty term proposed by Gulrajani\etal~\cite{gulrajani2017improved} for a more stable convergence. 
We set the batch size to $1$, and $\{\lambda_\text{VGG},\lambda_\text{GAN},\lambda_\text{CL},\lambda_\text{ID},\lambda_\text{GI},\lambda_\text{GL}\} = \{1.0,0.1,1.0,1.0,0.1,0.1\}$ in our experiments. Since our conditional label-to-face transformer only focuses on generating the foreground, we adopt an off-the-shelf image inpainting method~\cite{liu2018image} to complete the background, on which the generated foreground is pasted.

\begin{figure}[!t]
    \centering
    \includegraphics[width=0.88\linewidth]{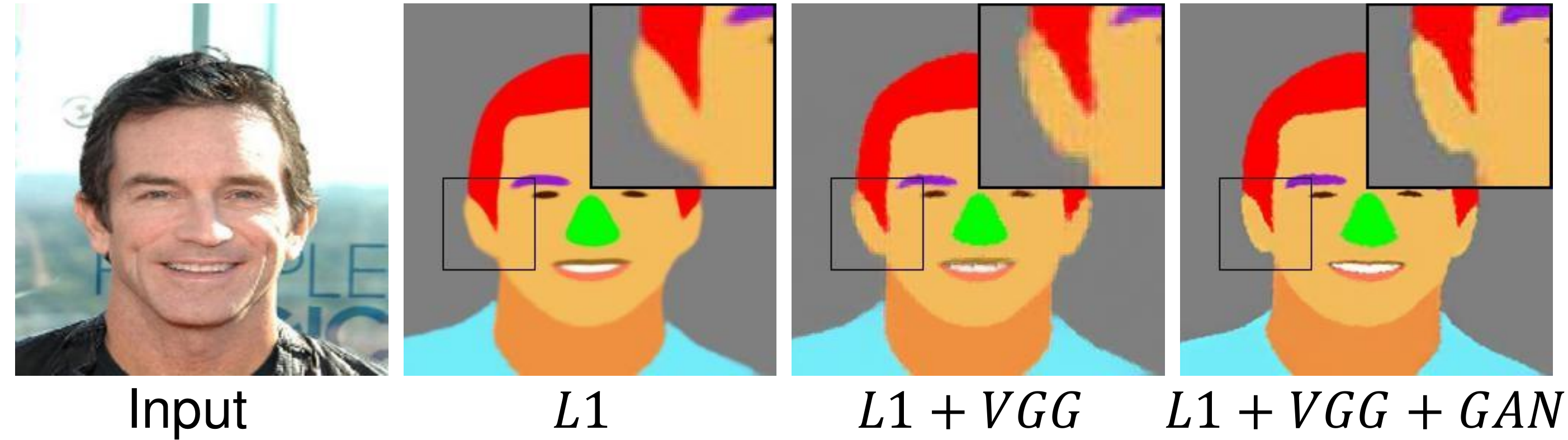}
    \caption{Different losses adopted for training the overall decoder. Merely using the $L_1$ loss creates blurry results. Introducing the VGG loss brings more details, but the results are not sharp enough. Adopting $L_1$+VGG+GAN achieves the most accurate label maps. }
    \label{fig:overall_decoder}
    \vspace{-5pt}
\end{figure}

\begin{figure}[!t]
    \centering
    \includegraphics[width=0.9\linewidth]{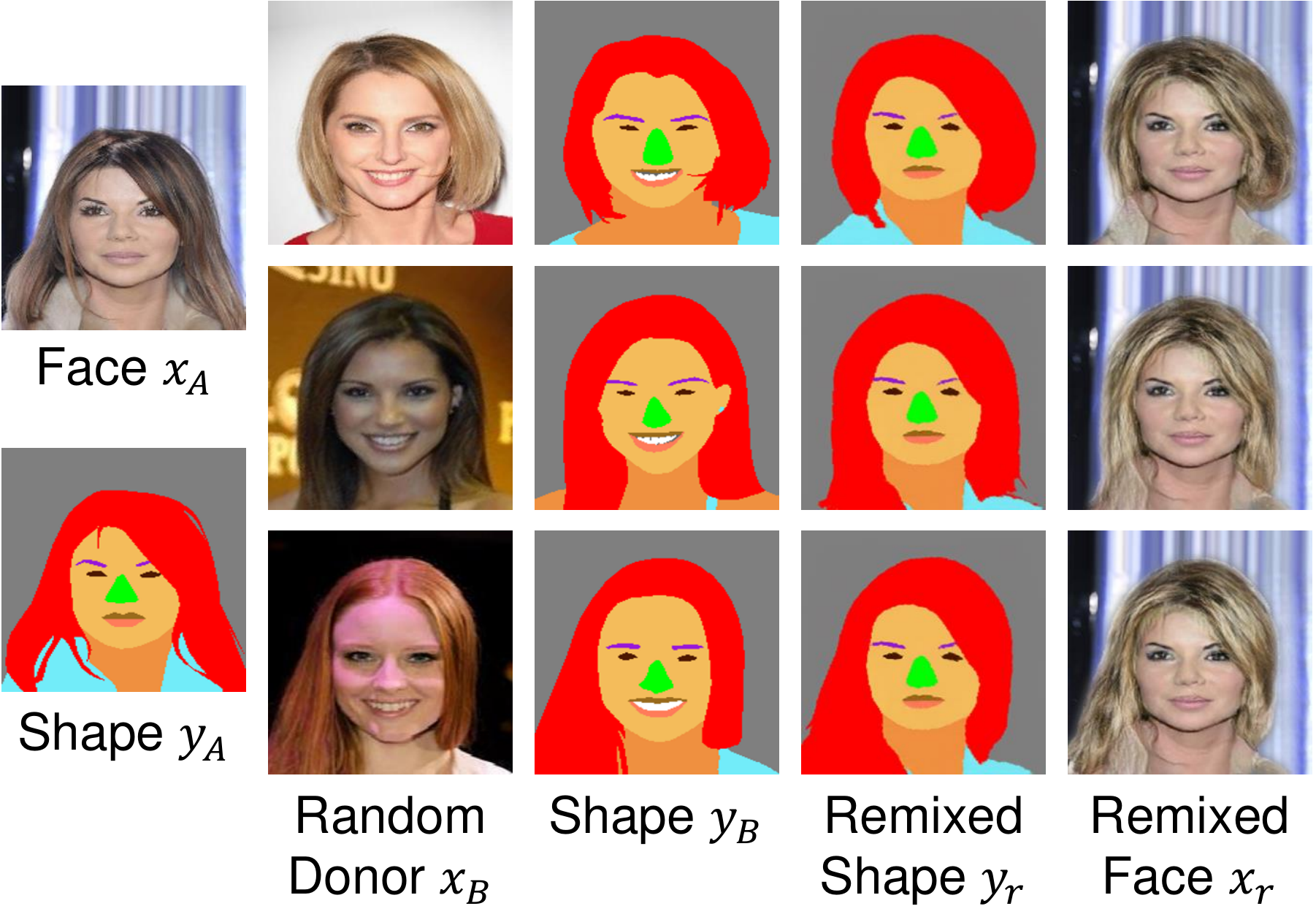}
    \caption{Different hair remixing results for a given face. We fix the receptor $x_A$ and show results with various donors $x_B$. }
    \label{fig:different_hair}
    \vspace{-5pt}
\end{figure}

\begin{figure}[!t]
    \centering
    \includegraphics[width=1.0\linewidth]{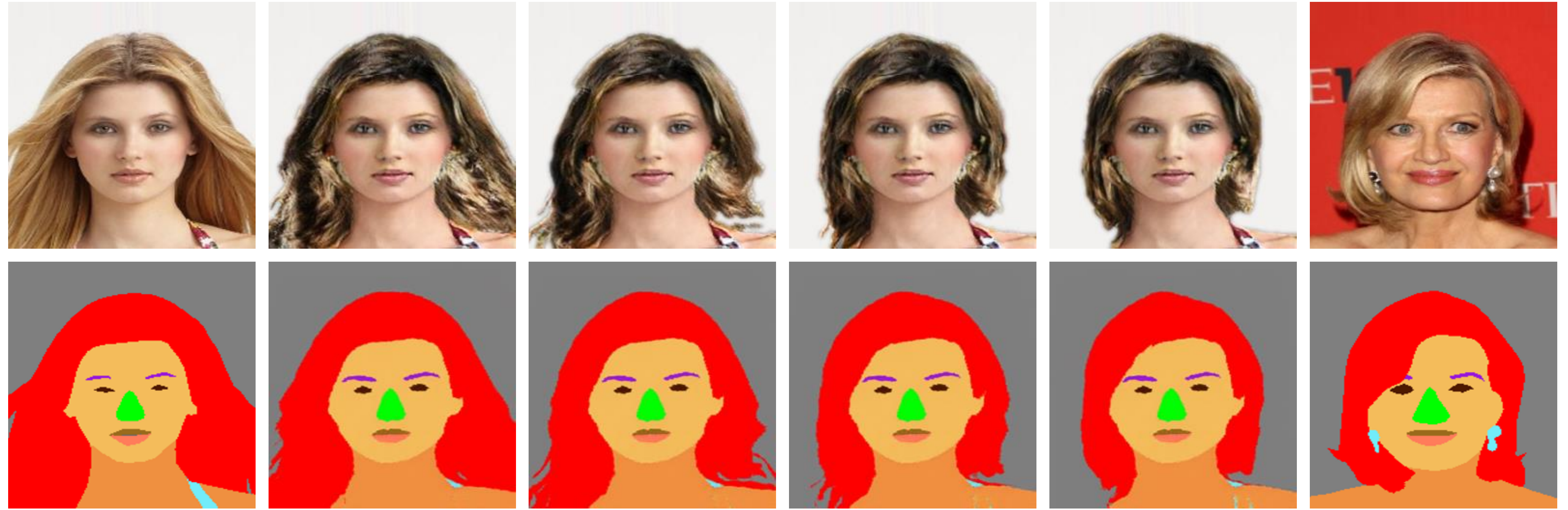}
    \caption{Part-wise feature interpolation. By interpolating the hair features between two images, we can create a series of intermediate results changing the hair style gradually.}
    \label{fig:interpolation}
    \vspace{-5pt}
\end{figure}

\begin{figure}[!t]
    \centering
    \includegraphics[width=1.0\linewidth]{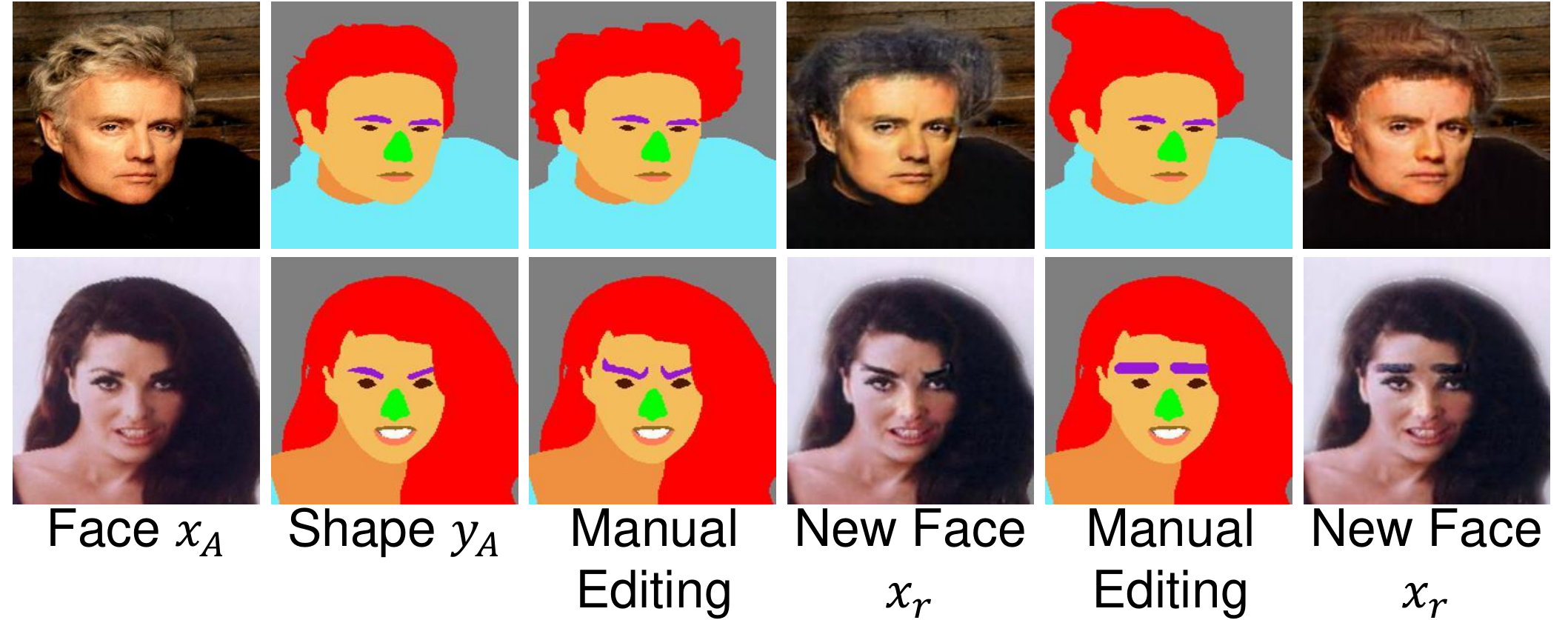}
    \caption{The conditional label-to-face transformer can also be applied to a manually edited label map. }
    \label{fig:manual_editing}
    \vspace{-5pt}
\end{figure}

\begin{figure*}[!t]
    \centering
    \includegraphics[width=0.80\linewidth]{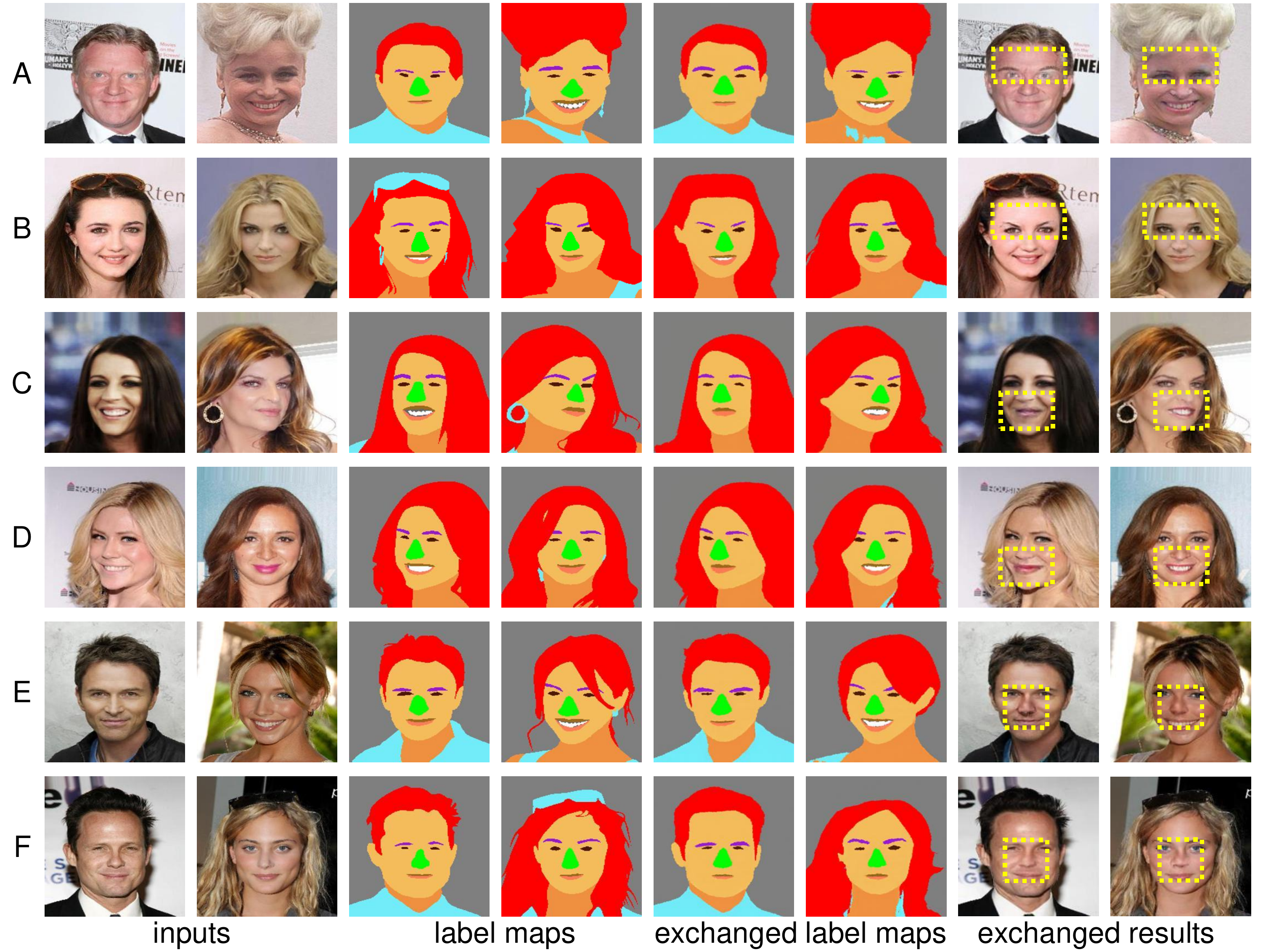}
    \caption{Exchanging different facial parts. Our system also supports exchanging the shapes between other facial parts. We exchange the eyebrows in A and B rows, the mouths in C and D rows, and the noses in E and F rows.}
    \label{fig:individual-exchanging}
    \vspace{-5pt}
\end{figure*}

\vspace{-2pt}
\subsection{Losses for Training The Overall Decoder}
We test three different loss settings for training the overall decoder, including $L_1$, $L_1$+VGG, and $L_1$+VGG+GAN. From Fig.~\ref{fig:overall_decoder} we can see that using merely an $L_1$ loss leads to blurry results. Introducing a VGG loss can bring more details. By combining the $L_1$, VGG and GAN losses together, we can enforce the overall decoder to produce high-quality label maps with finer details.

\vspace{-2pt}
\subsection{Part-Wise Shape Editing for Faces}
In this session we will show some part-wise shape editing manipulations supported by our proposed system. Firstly, as shown in Fig.~\ref{fig:different_hair}, we can easily transfer the hair style of a reference face $x_B$ to the input face $x_A$, while preserving the identity of $x_A$. 
Secondly, by interpolating the part-wise shape feature for the hair, we can gradually change the hair style of the target person from one to another, as shown in Fig.~\ref{fig:interpolation}. This indicates that our FaceShapeGene lies on a continuous manifold, where similar shapes have similar features. Note that the shapes of other parts keep unchanged throughout the whole interpolation process, which indicates that we successfully disentangle the shape features of different facial parts.
Thirdly, besides using the shape-remix network to provide a remixed label map, we can also manually edit a label map to flexibly control the shape of each facial part. Fig.~\ref{fig:manual_editing} demonstrates that we can generate a desired photo-realistic face image by conveniently editing the label map.

Other than the hair, we can also manipulate the features of other facial parts to accomplish partial face editing. Examples of exchanging the eyebrows, mouths, and noses between two faces are shown in Fig.~\ref{fig:individual-exchanging}.


\vspace{-2pt}
\subsection{Comparison to Existing Methods}
\begin{figure*}[!t]
    \centering
    \includegraphics[width=0.95\linewidth]{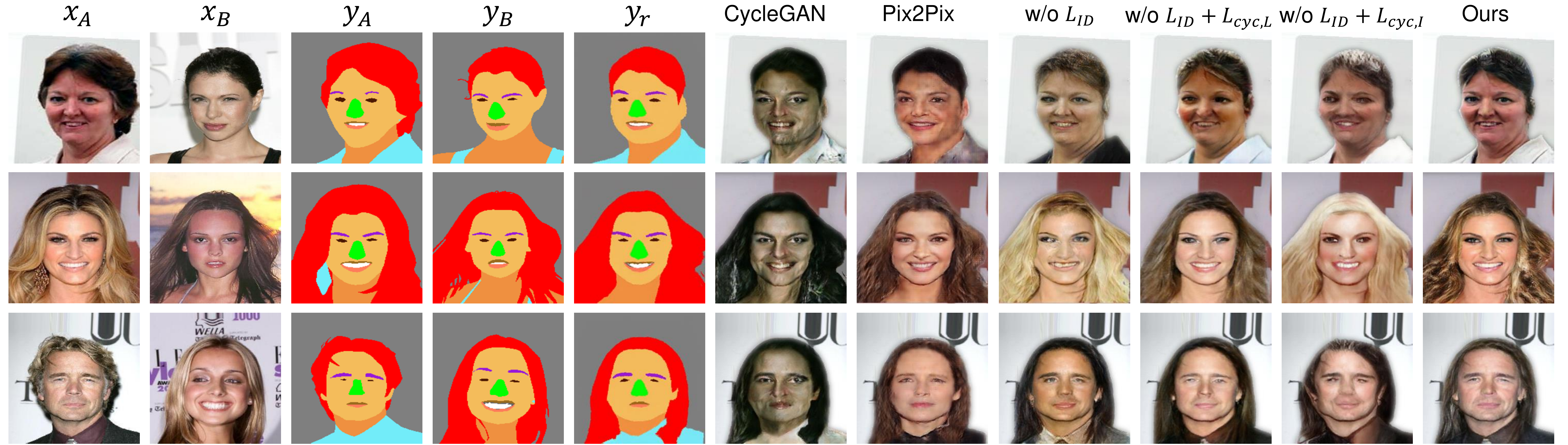}
    \caption{Comparison between different methods. While Pix2Pix and CycleGAN do not preserve the identity, our method can correctly conduct the label-to-face transformation and keep the identity of the receptor $x_A$. 
    The identity preservation ability is weakened when the identity and cycle-consistent constraints are removed.
    }
    \label{fig:ablation}
    \vspace{-5pt}
\end{figure*}

To our knowledge, our method is the first end-to-end system designed for automatically transferring the partial shape of one face to another. Thus, it is impossible to compare our full system to other methods. To demonstrate the superiority of the proposed training strategy for the conditional label-to-face transformer, we compare our method against two classic image-to-image translation methods Pix2Pix~\cite{isola2017image} and CycleGAN~\cite{zhu2017unpaired}. 
We also evaluate several variants of our proposed methods.

\vspace{-10pt}
\paragraph{Qualitative Comparisons.}
Fig.~\ref{fig:ablation} shows some visual results. Pix2Pix and CycleGAN cannot preserve the identity of the receptor $x_A$, while our method produces a photo-realistic remixed face keeping the identity of $x_A$. Removing $\mathcal{L}_\text{ID}$ impedes the identity preservation ability a little bit. The quality gets worse when $\mathcal{L}_\text{cyc,L}$ and $\mathcal{L}_\text{cyc,I}$ are also removed. Besides the desired part, other facial parts also change accordingly. Hence, the transformer fails to preserve the identity of the conditional input.

\vspace{-10pt}
\paragraph{Quantitative Comparisons.} To quantitatively compare our method with the CycleGAN and Pix2Pix baselines, and the variants of our method, we use Fr\'{e}chet Inception Distance (FID) score~\cite{heusel2017gans}, Inception Score (IS)~\cite{salimans2016improved}, and OpenFace Score~\cite{amos2016openface} as the evaluation metrics.
FID uses the Inception network~\cite{szegedy2015going} to extract features from an intermediate layer, and then evaluate the distance between the feature distributions of the ground-truth images and the generated images. A smaller FID score indicates more photo-realistic results. Inception Score is computed as the KL-divergence between the conditional class distribution and the marginal class distribution, where the output class label of the generated image is predicted by the Inception network. To evaluate the identity preservation ability, we exploit the OpenFace model~\cite{amos2016openface} to compute the face-id features of face images. The OpenFace score is computed as the dot product between the face-id features of the output face and the receptor face. 
Please refer to the supplementary material for the details of all the evaluation metrics.

We use all the images in the test set as receptors, while for each receptor we randomly choose a donor for it. The choice of the donor is fixed when testing different methods. The average scores over the test set are reported in Table~\ref{tab:ablation}. According to Table~\ref{tab:ablation}, our method produces the most photo-realistic results and has the best identity preservation ability. When $\mathcal{L}_\text{ID}$ is removed, the scores decline because the preservation of the identity is weakened. The scores gets even worse when $\mathcal{L}_\text{cyc,L}$ and $\mathcal{L}_\text{cyc,I}$ are also removed. We can also find that $\mathcal{L}_\text{cyc,I}$ has a greater impact on the scores than $\mathcal{L}_\text{cyc,L}$, because $\mathcal{L}_\text{cyc,I}$ directly constrains the reconstructed image while $\mathcal{L}_\text{cyc,L}$ only constrains the intermediate result.

\begin{table}[!t]
\centering
\resizebox{0.9\columnwidth}{!}{
\begin{tabular}{cccc}
\toprule
\hline
Method& FID Score& Inception Score& OpenFace \\
\midrule
\hline
CycleGAN& 45.07 & 2.296 & 1.947\\
Pix2Pix& 25.50 & 1.946 & 1.700\\
\emph{w/o} $\mathcal{L}_\text{ID}$& 18.55 & 1.959 & 1.322\\
\emph{w/o} $\mathcal{L}_\text{ID} + \mathcal{L}_\text{cyc,L}$& 18.15 & 2.214 & 1.252\\
\emph{w/o} $\mathcal{L}_\text{ID} + \mathcal{L}_\text{cyc,I}$& 21.97 & 2.726 & 1.538\\
Ours& \textbf{14.22}& \textbf{1.912} & \textbf{1.116} \\
\bottomrule
\hline
\end{tabular}
}
\caption{Quantitative Comparisons.}
\vspace{-5pt}
\label{tab:ablation}
\end{table}

\vspace{-10pt}
\paragraph{User Study.} We also conducted a user study to subjectively compare different methods. Our method was compared to CycleGAN, Pix2Pix, and ``\emph{w/o} $\mathcal{L}_\text{ID}$'', respectively. 
We randomly selected 10 groups of hair editing examples for testing. In each example, we showed 4 images, namely the receptor $x_A$, the donor $x_B$, our result, and the result of an alternative method.  On one hand, each participant was asked to choose a face image with a higher visual quality, considering both the overall fidelity and the success degree of changing the hair style. On the other hand, each participant was required to choose the result which preserves the identity better. 31 people participated in this study.
When compared to CycleGAN, 84\% of the participants thinks our synthesis quality is better, while 99\% thinks our method preserves the identity better. When compared to Pix2Pix, the numbers are 56\% and 93\%. When compared to ``\emph{w/o} $\mathcal{L}_\text{ID}$'', the numbers are 76\% and 97\%. The user study results show that our method performs better than other alternatives subjectively.


\vspace{-6pt}
\section{Conclusion} \label{sec:conclusion}
\vspace{-2pt}
In this paper, we proposed \emph{FaceShapeGene}, a disentangled shape representation for face, which encodes the shape information of each facial part separately.
Exploiting the \emph{FaceShapeGene}, we developed a novel face editing system, which includes a shape-remix network and a conditional label-to-face transformer. A cyclic training strategy was further proposed to train the system in an unsupervised manner. The extensive experiments demonstrate that our system can achieve state-of-the-art partial editing results.

\clearpage

{\small
\bibliographystyle{ieee}
\bibliography{egpaper_for_review}
}

\end{document}